# An End-to-End Multi-Scale Network for Action Prediction in Videos

Xiaofa Liu, Jianqin Yin, *Member, IEEE,* Yuan Sun, Zhicheng Zhang, Jin Tang

*Abstract*—In this paper, we develop an efficient multi-scale network to predict action classes in partial videos in an end-to-end manner. Unlike most existing methods with offline feature generation, our method directly takes frames as input and further models motion evolution on two different temporal scales. Therefore, we solve the complexity problems of the two stages of modeling and the problem of insufficient temporal and spatial information of a single scale. Our proposed End-to-End Multi-Scale Network (E2EMSNet) is composed of two scales which are named segment scale and observed global scale. The segment scale leverages temporal difference over consecutive frames for finer motion patterns by supplying 2D convolutions. For observed global scale, a Long Short-Term Memory (LSTM) is incorporated to capture motion features of observed frames. Our model provides a simple and efficient modeling framework with a small computational cost. Our E2EMSNet is evaluated on three challenging datasets: BIT, HMDB51, and UCF101. The extensive experiments demonstrate the effective-ness of our method for action prediction in videos.

## I. INTRODUCTION

THE goal of action prediction in videos is to predict the class label of an ongoing action from an observed part of it over temporal axis so far[1]. It is a subset of a broader research domain on human activity analysis. Different from conventional action recognition with fully executed actions[2][3][4], it is more challenging to predict the action label in ongoing actions due to the incompleteness of actions and the continuous evolution of actions. It has attracted a lot of research attention because of its wide application in some scenarios with high real-time requirements, such as human-machine interaction, security surveillance, etc.

Although the previous work has achieved promising results by adopting a two-stage approach, there generally had problems of complex modeling and feature redundancy. The previous method separated feature extraction from predictive modeling[5][6][7][8][9][10][11][12]. This separation operati-on makes the spatio-temporal representation obtained may deviate from the action prediction. Moreover, it complicates the model design. Secondly, because the feature is generated offline, the complete action must be divided into fixed segments in advance, which not only results in the redundancy of the feature in the time dimension, but also is not applicable to the evolving action.

Therefore, in this paper, we propose an end-to-end method, which effectively reduces the complexity of the model and introduces more fine-grained spatio-temporal information. We designed the end-to-end network from three aspects, sampling method, local spatio-temporal information representation, and long-term time sequence fusion. In order to adapt the end-to-end structure to the evolving motion, we first changed the preprocessing and feature generation method, which will be described in Part 3. Second, to reduce computational consumption to achieve end-to-end structure, we use 2D convolution instead of two-stream networks or 3D convolutions to extract local spatio-temporal features. Finally, to enhance the temporal information of action evolution, we present an observed global scale to fuse the historical evolution information of actions.

Similar to the application of spatial multi-scale in image field, multi-scale research in the temporal dimension is also increasing in video analytics. Compared to images, the variation of temporal scales in videos poses additional challenges. How to effectively utilize the motion evolution information at different time scales has gradually gained attention in video motion analysis. Feichtenhofer[4] et al. proposed SlowFast network for video recognition. Their method utilizes two branches, a slow pathway with low frame rate and a fast pathway with high frame rate, to capture spatial semantics and motion at fine temporal resolution. Wang[13] et al. proposed an efficient multi-scale model for action recognition, which utilizes short-term and long-term temporal difference modules to capture both short-term and long-term motion information better.

Most of the existing action prediction methods are insufficient to focus on multi-scale temporal, making them fail to capture fine-grained temporal information. They use a fixed frame rate to sample each partial video, and use a fixed temporal scale for feature generation and modeling[1][5][6][7][8][9][11]. Although these methods simplify the

▪ This work was supported partly by the National Natural Science Foundation of China (Grant No. 62173045, 61673192), partly by the Fundamental Research Funds for the Central Universities (Grant No. 2020XD-A04-3), and the Natural Science Foundation of Hainan Province (Grant No. 622RC675). *(Corresponding author: Jianqin Yin).*
▪ Xiaofa Liu is with the School of Modern Post, Beijing University of Posts and Telecom-munications, Beijing 100876, China (e-mail: liuxiaofamail@163.com )
▪ Jianqin Yin, Zhicheng Zhang, and Jin Tang are with the school of Artificial Intelligence, Beijing University of Posts and Telecommunications, Beijing 100876, China (e-mail: jqyin@bupt.edu.cn, zczhang@bupt.edu.cn, tangjin@bupt.edu.cn ).
▪ Yuan Sun is with Electronic Engineering School, Beijing University of Posts and Telecommunications, Beijing 100876, China (e-mail: sunyuan@bupt.edu.cn ).

processing of the input of feature generation and reduce the computation to a certain extent, they ignore the evolution of action. Too much fine-grained information will be lost, and the spatio-temporal information in the video cannot be fully utilized.

Our method takes both the local evolution information between adjacent frames and the global evolution information of the entire observed video sequence into account. Therefore, we design two temporal scales to increase fine-grained timing information. Firstly, the segment scale uses RGB frames with temporal difference to capture temporal information in each segment. Secondly, the observed global scale uses LSTM module to fuse all the observed action evolution information. Through modeling in short-term and long-term time scales, our method can be mining more fine-grained temporal information without increasing the computational load.

Our E2EMSNet provides a simple yet effective framework for the problem of ongoing action prediction in videos. In summary, our main contributions lie in the following three aspects:

● We propose a simple end-to-end approach for action prediction in videos. To the best of our knowledge, this is the first work focusing on this problem.

● We investigate two scales in the temporal dimension to model the evolution of actions, and propose a segment summarization and propagation framework. The segment scale is used to model the local evolution of the action, and the observed global scale is used to model the global evolution of the action.

● We achieve a trade-off of efficiency and effectiveness. We achieve state-of-the-art performance on several datasets while using only 2D convolutions framework and RGB format of features.

## II. RELATED WORK

### A. Action Recognition

Action recognition methods take fully observed videos as input and output labels of human actions. Action recognition has been extensively studied in past few years[2][3][4][13][14]. These studies can be roughly divided into two categories. Methods in the first category are two-stream CNNs, which was first proposed in[15]. It used two inputs of RGB and optical flow to model appearance and motion information separately in videos with a late fusion. In addition, follow-up research has adopted two RGB inputs sampled at different FPS or carefully designed temporal modules for efficiency, including Non-local Net[16], STM[17], SlowFast[4], and Correlation Net[18]. The second method is to use 3D CNNs[19][20]. It proposed 3D convolution and pooling to learn spatiotemporal features from videos directly. Several variants adopted a 2D + 1D paradigm to reduce the computation cost of 3D convolution, which implement by decomposing 3D CNNs into a 2D convolution and a 1D temporal convolution[21][22][23]. Several works focused on designing more powerful and efficient temporal modules, such as TSM[14], TAM[24], TEA[25], and TDN[13]. More recent works tried clip-based architecture search for video recognition, focusing on capturing appearance and motion or context information in a more fine-grained and efficient manner[13][26]. Although these methods mainly learned features for the videos with full action executions, their core ideas have certain reference significance for ongoing action prediction in videos.

### B. Action Prediction

Action prediction methods were proposed to predict the action given a partially observed video. [9] was the first work along these lines, they formulated the problem probabilistically and proposed a dynamic bag-of-words approach, modeling how feature distributions of activities change as observations increase. In the last decade, researchers approach this task from various perspectives and can be grouped into three major divisions[27]. The first method can be formulated as one-shot mappings from partial observations to groundtruth labels of full observations. The basic assumption underlying these methods is that a partial observation of an action video provides sufficient information to define the appropriate overall action class regardless of the unobserved part. Follow-up research work[28][29][6][30] adopted more robust features, hierarchical extractions, and learning-based classifiers to perform more fine-grained analysis of an initial partial observation for better performance. The second division is knowledge distillation-based methods. These methods distill the information from the full observations into partial observations[31][5][11][32]. These methods attempted to lend power from unobserved data in training to either enrich the feature representation of partial data or encourage the classifiers to easily recognize partial data. Another way to exploit future information is by propagating the partial observation into the future in a temporal extrapolation fashion[33][34][12][35][36]. For example, [12] learned to propagate frame-wise residuals in feature space to complete partial observation.

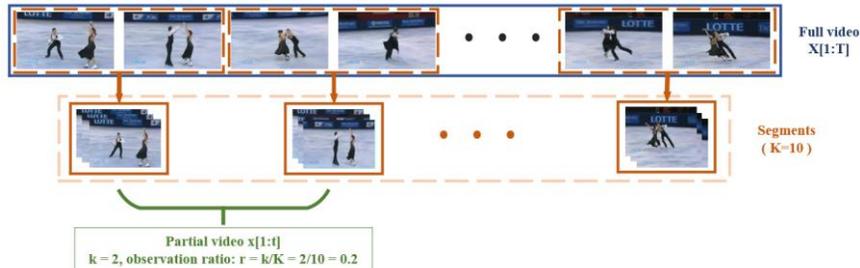

**Fig. 1.** Relevant definitions in action prediction in videos: full video, partial video, segments, and observation ratio.

*C. Multiple temporal scales for action analysis in videos*

Temporal sequence forecasting usually faces the following situations for scenarios with insignificant periodic motion: long-term forecasts need to consider trend information (long-term dependencies), and short-term forecasts need to consider fine-grained volatility (short-term dependencies). The current difficulty is how to model long-term dynamic dependencies and consider long-term and short-term dependencies. There are two methods currently. The main existing method is hierarchical modeling, which is achieved by establishing hidden layers of different granularities[37][38][39][40][41] or decomposing the original data to obtain data of different granularities[42][43]. The second method is designing the gate mechanism, which achieved by modifying the internal structure of RNN[44]. We inherit this idea that both long-term and short-term dependencies in video must be carefully considered, and a trade-off approach is adopted.

## III. OUR METHOD

In this section, we detail our approach to mining ongoing action evolution information in videos using multiple scales in an end-to-end fashion. Specifically, we first describe the problem formulation. Then, we elaborate on our end-to-end framework and method for multi-scale modeling of ongoing action sequences.

*A. Problem formulation*

Given a video containing human motion (the video may contain arbitrary incomplete motion), the goal is to predict the class label. We follow the problem formulation in the[31], which has been widely adopted in subsequent work[5][7][11]. As shown in Fig. 1, Given a *full video* $X[1:T]$ with complete action execution, 1 represents the first frame of the video, and $T$ represents the last frame. We use $x[1,t], t \in [1,T]$ to simulate the action execution in video from 1 to $t$, defined as *partial video*. In order to facilitate quantitative experiments, we usually divide a full video into $K$ segments, each containing $(T/K)$ frames. Assuming that the action is executed to the $kth, k = [1,2,...,K]$ segment, the *observation ratio* is defined as $r = k/K$. As defined above, as shown in Fig.1, the full video $X$, is divided into $K$ segments. Among them, the partial video marked with green has an observation ratio $r = k/K = 2/10 = 0.2$, and it can be considered that its action has been executed 20%.

*B. Data processing*

We adopt a data processing method different from the previous method. As shown in Fig. 2, the upper part is the data processing method used in the previous method. They first divided a complete video $X$ into $K$ segments, and combined segments into partial videos to simulate action evolution. Then the partial video is sampled to extract the spatio-temporal representation. The problem caused by this is that each partial video needs to be separately extracted for spatio-temporal representation, which divides the continuous evolution of action. The feature extraction of partial videos with higher observation rates cannot use the previous partial videos with lower observation rates. It will cause redundancy in the time dimension. At the same time, with the increase in the observation rate, the temporal information will become more and more sparse. Compared with them, we directly extract the local spatio-temporal representations of each segment. In this way, the previous spatio-temporal information can be continuously used with the evolution of actions. This makes our model more robust to action duration, and more abundant spatio-temporal information can be obtained.

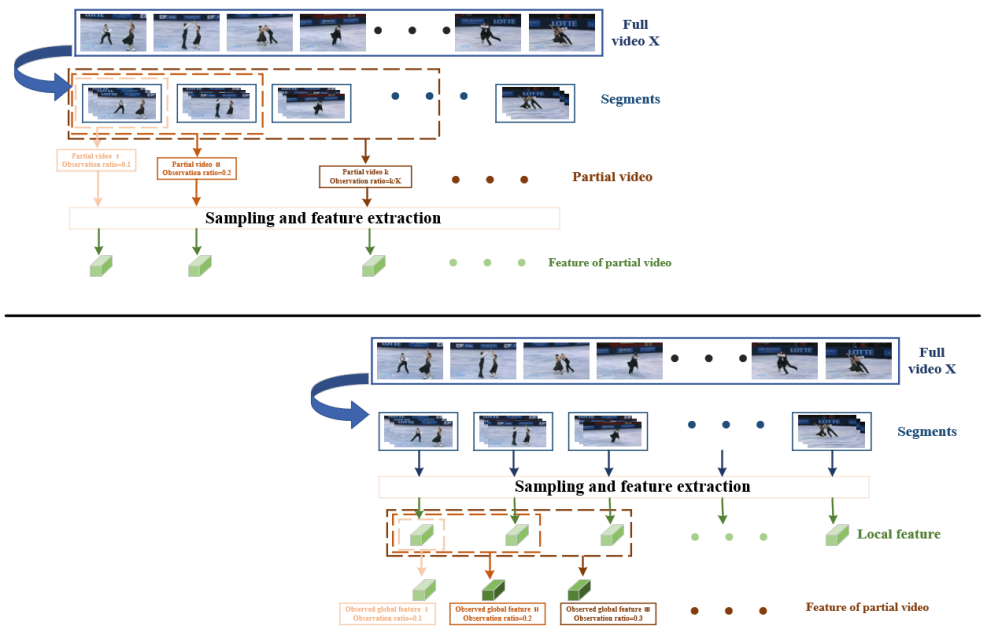

**Fig. 2.** Differences in data processing between our method and previous methods. The upper is the data processing method used in the previous method, and the lower is the data processing strategy used in our method.

*C. Network architectures*

In this subsection, we elaborate on our network structure. Due to the data processing method mentioned in the previous section and the design of network structure, we can model action evolution in a finer-grained manner without increasing the computational load. First, we introduce how to extract short-term features for short time windows, which we call the segment scale. Then, we introduce how to fuse the segment scale to generate observed global features for the observed local videos.

**Segment scale.** Compared with images, video is a dynamic sequence of pictures arranged in time, so the temporal context relationship of frames and the spatial relationship organization of a single frame need to be considered simultaneously. For extracting and fusion of two kinds of relations in local time windows, directly stacking frames as input will bring a lot of redundant information. This method is inefficient. Moreover, it will introduce too much noise and reduce the robustness of the model. If only a single image frame is used as input, the dynamic information of the temporal window will be lost. RGB temporal difference turned out to be an efficient alternative modality to optical flow as motion representation [45][13]. To extract the spatio-temporal features of each local temporal window, we adopt the idea in[13] as a short-term feature extraction module. Different from action recognition, in the action prediction problem, we cannot get the spatio-temporal information after the current frame, so we only keep the short-term TDM (temporal difference module) in[13]. Specifically, for each segment, we randomly sample 5 frames $I = [I_{t-2}, I_{t-1}, I_t, I_{t+1}, I_{t+2}]$, then the RGB difference information of these frames is down-sampled, and the 2D convolutions network is used to obtain the depth feature $S(I_i)$, as expressed in Equation (1).

$$S(I_i) = Upsample(CNN(Downsample(D(I_i)))) \quad (1)$$

At the same time, to preserve the original frame-level representation as much as possible, we fuse the original features $I_t$ with $S(I_i)$ after convolutions (in our actual experiment, the original feature passes through a layer of 2D CNN, as shown in Equation (2)).

$$S(fuse) = S(I_i) + CNN(I_t) \quad (2)$$

The fused feature is fused again with the feature from RGB difference (Equation (3)). Finally, the feature of each segment is obtained, which is the representation of segment scale.

$$S(out) = CNN(S(fuse)) + CNN(Downsample(D(I_i))) \quad (3)$$

**Observed global scale.** In action prediction, the action evolution of the human body is an ongoing sequence of information, and we use the observation rate to simulate its progress. Therefore, the segments are temporally sequential, and the representative actions can only evolve from front to back. In the previous section, we model the local spatio-temporal action of each segment. More logically, as time progresses, each segment's local temporal window is added to the historical sequence before it. Therefore, the crux of the problem is how to effectively utilize all observed segments to reconstruct the historical global evolution.

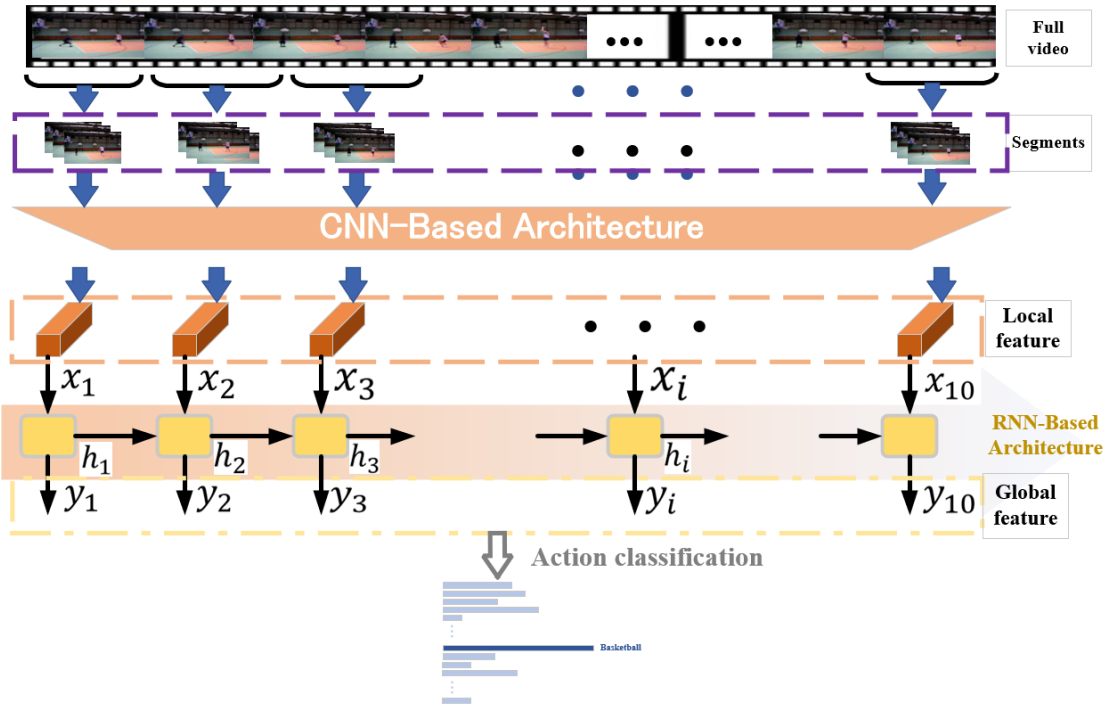

**Fig. 3.** Overview of End-to-End Multi-scale Network. Given a full video, split it into K segments. For each segment, a CNN-based module extracts the local motion evolution to achieve more fine-grained modeling, which we call the segment scale. Then, temporal modeling is performed on each segment in chronological order to model the observed global action evolution, which we call the observed global scale.

Moreover, in the actual scene, the evolution of the action cannot know its end time and duration, which means that the overall length of the history is uncertain. Therefore, it is natural to use the variable-length input characteristics of LSTM to model the global spatiotemporal characteristics of historical observations, as shown in formula (4).

$$Y(i) = L(S(out)) \quad (4)$$

As shown in Fig. 3, when the action evolves to the third segment, the LSTM adds the short-term time window of the third segment to the historical observation in the time dimension. Implemented the observed global evolution to model the first three segments progressively. In this way, the spatiotemporal relationship in each segment can be modeled in a more fine-grained manner, and the subsequent segments are modeled in a progressive manner to model the historical global history without additional computational consumption.

## IV. EXPERIMENTS

In this section, we present the experiment results of our framework. First, we describe the evaluation datasets and implementation details. Then, we compare our E2EMSNet with state-of-the-art methods.

### A. Datasets

We evaluate our method on three video datasets: BIT[46], HMDB51[47] and UCF101[48]. **BIT** consists of 8 classes of human interactions (bow, boxing, handshake, high-five, hug, kick, pat, push), with 50 videos per class. Videos are captured in realistic scenes with cluttered backgrounds, partially occluded body parts, moving objects, and variations in subject appearance, scale, illumination condition, and viewpoint. Even though BIT has a limited number of classes and videos, it is a complex dataset because of their backgrounds and the similarity of the beginning and ending scenes. The ratio of videos between training and testing is 17:8. **HMDB51** is a large-scale human action recognition dataset that comprises 51 daily action categories. It contains some fine-grained human facial motions, such as smiling, laughing, etc, in static background windows, which are not seen in other comparable datasets, and challenges the spatiotemporal modeling of actions. There are 6766 video clips with at least 102 videos for each class. There are three official data splits. **UCF101** is a dataset collected from Youtube and trimmed for action recognition (each video contains exactly one action). It includes 101 distinct action classes and 13320 overall video clips with at least 100 videos for each category. All videos are divided into 25 groups and updated with the setup of Three Train/Test Splits.

### B. Implementation details

Thanks to our end-to-end network structure design, we can easily generalize to various video datasets. In experiments, we use ResNet50 with the short-term module in [13] to build segment scale. On the three datasets, we simulated the action evolution with the observation rate from 0.1 to 1, with a step size of 0.1, to obtain ten segments, and use each segment as a segment scale. Our network structure can use any length and number of segments as the segment scale. For each segment, we randomly sample 5 frames for computing RGB differential information. We employ convolutional layers pre-trained on kinetics400, and set dropout to reduce overfitting. We first convert the video into video frames, and each video frame is resized to have shorter side in [256, 320] and a crop of 224×224 is randomly cropped. We use two NVIDIA GeForce RTX 3090s to train our model. On the BIT dataset, we follow the official settings to divide the training set and test set. Specifically, in each category, 34 videos are used as the training set, and 16 videos are used as the test set. On the HMDB51 dataset, we follow the standard evaluation protocol using three training/testing splits, and report the average accuracy over three splits. On the UCF101 dataset, we use the first 15 groups of videos for model training, the following 3 groups for model validation, and the remaining 7 groups for testing.

### C. Comparison with the state of the art

In this subsection, we compare out E2EMSNet with those state-of-the-art methods, including DBoW[9], MTSSVM[28], MMAPM[31], Deep-SCN[5], AAPNet [49], RGN-KF[12], RSPG + AS-GCN[8], AORAP[50], and AASE +JOLO-GCN[51] on the BIT dataset, MTSSVM[28], Global-local[52], AKT[7], STRR[30] on the HMDB51 dataset, MTSSVM[28], DeepSCN[5], AAPNet[49], Teacher-Student[11], RGN-KF [12], RSPG + AS-GCN[8], SPR-Net[53], JVS + JCC + JFIP[32], STRR (ResNet18) [30], and Xinxiao Wu et al.[54] on the UCF101 dataset. We reported the results of these compared methods provided by authors.

Table Ⅰ illustrates the accuracy of action prediction and compares our method with several state-of-the-art methods on the BIT dataset. As seen from the results, our method achieves significant improvements in observation rates from 0.1 to 1. This can be explained by the fact that our method can make reliable predictions on actions as the actions evolve.

TABLE I

THE ACCURACY (%) OF DIFFERENT ACTION PREDICTION METHODS ON BIT DATASET AT DIFFERENT OBSERVATION RATIOS FROM 0.1 TO 1. NOTE THAT THE MISSING VALUE IS BECAUSE THE EXPERIMENTAL RESULTS OF THE CORRESPONDING OBSERVATION RATE ARE NOT PROVIDED IN THE ORIGINAL PAPER.

| Method | Input | Feature-dim | Observation Ratio | | | | | | | | | | |
|---|---|---|---|---|---|---|---|---|---|---|---|---|---|
| | | | 0.1 | 0.2 | 0.3 | 0.4 | 0.5 | 0.6 | 0.7 | 0.8 | 0.9 | 1.0 | Avg. |
| DBoW[9] | | Hand-crafted | 22.66 | 25.78 | 40.63 | 43.75 | 46.88 | 54.69 | 55.47 | 54.69 | 55.47 | 53.13 | 45.31 |

| Method | Input | Feature-dim | 0.1 | 0.2 | 0.3 | 0.4 | 0.5 | 0.6 | 0.7 | 0.8 | 0.9 | 1.0 | Avg. |
|---|---|---|---|---|---|---|---|---|---|---|---|---|---|
| MTSSVM[28] | | Hand-crafted | 28.12 | 32.81 | 45.31 | 55.45 | 60.00 | 61.72 | 67.19 | 70.31 | 71.09 | 76.56 | 56.85 |
| MMAPM[31] | | Hand-crafted | 32.81 | 36.72 | 53.90 | 59.38 | 67.97 | 63.28 | 68.75 | 75.00 | 75.78 | 79.90 | 61.32 |
| DeepSCN[5] | RGB | 3D-CNN+Hand-crafted | 37.50 | 44.53 | 59.83 | 71.88 | 78.13 | 85.16 | 86.72 | 87.50 | 88.28 | 90.63 | 73.01 |
| AAPNet[49] | RGB | 3D-CNN+Hand-crafted | 38.84 | 45.31 | 64.84 | 73.40 | 80.47 | 88.28 | 88.28 | 89.06 | 89.84 | 91.40 | 74.97 |
| RGN-KF[12] | RGB+Flow | 2D-CNN | 35.16 | 46.09 | 67.97 | 75.78 | 82.03 | 88.28 | 92.19 | 92.28 | 92.16 | 92.16 | 76.41 |
| RSPG+AS-GCN[8] | Skeleton | LSTM | 55.70 | | 77.30 | | 91.00 | | 93.00 | | 93.00 | 94.00 | |
| AORAP[50] | RGB+Flow | 2D-CNN | 40.16 | | 71.48 | | 92.89 | | 96.8 | | | 96.48 | 79.56 |
| AASE + JOLO-GCN[51] | Skeleton | LSTM | | | 80.20 | | 92.40 | | | | | | |
| OCRL [6] | RGB | 3D-CNN | | | 65.6 | | 84.4 | | 90.6 | | 89.1 | | |
| **E2EMSNet (Ours)** | RGB | 2D-CNN+LSTM | **82.81** | **89.06** | **96.88** | **98.43** | **98.43** | **96.88** | **100** | **100** | **100** | **100** | **96.25** |

Table II shows the experimental results on the HMDB51 dataset, and table III shows the experimental results on the UCF101 dataset. Thanks to the design of our segment scale, action evolution can be modeled in a more fine-grained way. As shown in the table, at 0.2 of observation rate, the accuracy rate on HMDB51 dataset is increased by more than 10%, and the accuracy rate on UCF101 in increased by more than 3% except the results in[32]. This means that our method can better predict its class in the early stages of the action. As the observation rate increases, our method can achieve a more competitive performance, although the performance improvement is limited.

At the same time, we have to admit that on the HMDB51 and UCF101 datasets, although our method has achieved relatively good performance when the observation rate is low, as the action continues to evolve and the temporal scale continues to grow, our model is limited in the later observation ratios. We think that the modeling ability of observed global scale for long time windows is insufficient.

TABLE II

THE ACCURACY (%) OF DIFFERENT ACTION PREDICTION METHODS ON HMDB51 DATASET AT DIFFERENT OBSERVATION RATIOS FROM 0.1 TO 1. NOTE THAT THE MISSING VALUE IS BECAUSE THE EXPERIMENTAL RESULTS OF THE CORRESPONDING OBSERVATION RATE ARE NOT PROVIDED IN THE ORIGINAL PAPER.

| Method | Input | Feature-dim | 0.1 | 0.2 | 0.3 | 0.4 | 0.5 | 0.6 | 0.7 | 0.8 | 0.9 | 1.0 | Avg. |
|---|---|---|---|---|---|---|---|---|---|---|---|---|---|
| MTSSVM[28] | | Hand-crafted | 13.60 | | 26.70 | | 33.80 | | 37.80 | | 38.80 | | |
| Global-local[52] | | Hand-crafted | 38.80 | 43.80 | 49.10 | 50.40 | 52.60 | 54.70 | 56.30 | 56.90 | 57.30 | 57.30 | 51.72 |
| AKT[7] | RGB | 3D-CNN | 43.50 | 48.40 | 51.20 | 54.20 | 56.40 | 58.40 | 59.60 | 60.20 | 61.10 | 61.80 | 55.48 |
| STRR[30] | RGB | 3D-CNN | 45.10 | | 52.35 | | 56.73 | | 5941 | | 61.11 | | |
| **E2EMSNet (Ours)** | RGB | 2D-CNN+LSTM | **59.21** | **60.52** | **62.23** | **64.47** | **64.73** | **64.86** | **64.86** | **65.26** | **65.13** | **65.39** | **63.67** |

Table III

THE ACCURACY (%) OF DIFFERENT ACTION PREDICTION METHODS ON UCF101 DATASET AT DIFFERENT OBSERVATION RATIOS FROM 0.1 TO 1. NOTE THAT THE MISSING VALUE IS BECAUSE THE EXPERIMENTAL RESULTS OF THE CORRESPONDING OBSERVATION RATE ARE NOT PROVIDED IN THE ORIGINAL PAPER.

| Method | Input | Feature-dim | 0.1 | 0.2 | 0.3 | 0.4 | 0.5 | 0.6 | 0.7 | 0.8 | 0.9 | 1.0 | Avg. |
|---|---|---|---|---|---|---|---|---|---|---|---|---|---|
| MTSSVM[28] | | Hand-crafted | 40.05 | 72.83 | 80.02 | 82.18 | 82.39 | 83.12 | 83.37 | 83.51 | 83.69 | 82.82 | 77.39 |
| DeepSCN[5] | RGB | 3D-CNN+Hand-crafted | 45.02 | 77.64 | 82.95 | 85.36 | 85.75 | 86.70 | 87.10 | 87.42 | 87.50 | 87.63 | 81.30 |

| Method | Input | Architecture | 0.1 | 0.2 | 0.3 | 0.4 | 0.5 | 0.6 | 0.7 | 0.8 | 0.9 | 1.0 | Avg. |
|---|---|---|---|---|---|---|---|---|---|---|---|---|---|
| AAPNet[49] | RGB | 3D-CNN + Hand-crafted | 59.85 | 80.85 | 86.78 | 86.47 | 86.94 | 88.34 | 88.34 | 89.85 | 90.85 | 91.99 | 85.02 |
| Teacher-Student[11] | RGB | 3D-CNN | 83.32 | 87.13 | 88.92 | 89.82 | 90.85 | 91.04 | 91.28 | 91.23 | 91.31 | 91.47 | 89.63 |
| RGN-KF[12] | RGB + Flow | 2D-CNN | 83.12 | 85.16 | 88.44 | 90.78 | 91.42 | 92.03 | 92.00 | 93.19 | 93.13 | 93.13 | 90.24 |
| RSPG+AS-GCN[8] | Skeleton | LSTM | | | | 90.30 | **93.10** | | | | **94.70** | | |
| SPR-Net[53] | RGB | 3D-CNN | 88.70 | | | | 91.60 | | | | 91.40 | | |
| JVS+JCC+JFIP[32] | RGB | (2D+1D)-CNN | | | 91.70 | | | | | | | | |
| STRR (ResNet18)[30] | RGB | 3D-CNN | 80.86 | | 88.61 | | 89.31 | | 90.31 | | 89.82 | | |
| Xinxiao Wu et al.[54] | RGB + Flow | 2D-CNN | 82.36 | 85.57 | 88.97 | | 91.32 | | 92.41 | | **93.02** | | |
| **E2EMSNet (Ours)** | RGB | 2D-CNN + LSTM | **88.77** | 90.31 | **90.94** | **91.33** | 91.96 | **92.73** | **93.11** | **92.98** | 92.98 | 92.73 | **91.78** |

## D. Ablation study

Here, we provide more evaluation results on the UCF101 dataset.

**Influence of multi-scale architecture.** Table IV. Illustrates the results of the ablation study for different scale architecture. First, we introduce the details of the ablation study. Then, we analyze the effects of multi-scale architecture by comparing the results with different settings.

TABLE IV
THE ACCURACY (%) AT DIFFERENT SCALE SETTINGS ON THE UCF101 DATASET.

| Observation ratio | 0.1 | 0.3 | 0.5 | 0.9 | Avg. |
|---|---|---|---|---|---|
| The segment scale only | 90.56 | 91.58 | 91.83 | 91.45 | 91.55 |
| The segment scale+observed global scale | 90.05 | 90.82 | 92.60 | 92.47 | 91.78 |

'The segment scale only' uses the CNN-based module for action prediction. 'The segment scale + observed global scale' uses the CNN-based and LSTM modules to learn different scale information. In the first setting, for action clips with different observation rates, we sample 5 frames and use the segment scale only for prediction. In the second setting, we adopt a complete structure with segment scale and observed global scale. Even though the average accuracy difference is insignificant, the multi-scale structure is essential for ongoing action prediction. Results of 'The segment scale only' has little discrimination under different observation rates, as shown in Fig 4. This indicates that its feature representation and discriminative degree for different observation rates are insufficient. At the same time, due to the sparse sampling of long-time scales, we believe this manner will perform worse for complex actions and actions with long duration. Conversely, adding observed global scale and changing the sampling strategy will make the prediction process more cognitive (As the observation rate increases, the confidence of the prediction should be increasing.). Moreover, due to the more fine-grained feature extraction for actions, it has better robustness to complex and long-duration actions.

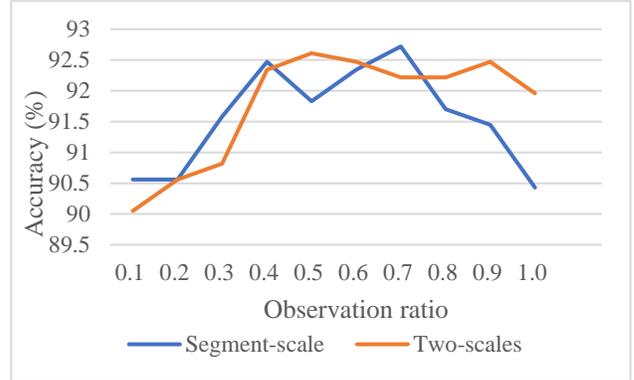

**Fig. 4.** Prediction accuracy (%) under two scale settings on UCF101 dataset.

**Influence of hyperparameters.** Finally, we briefly introduce the experimental results on UCF101 dataset under different hyperparameter settings. To ensure a single variable, we have conducted comparative experiments on the following hyperparameters, and the results are shown in Table V.

## E. Analysis of the performance of different actions

We follow the grouping of the UCF101 dataset and divide it into five groups: Human-Object interaction, Body-Motion only, Human-Human interaction, Playing musical instruments, and Sports. We selected three action categories under each group, for a total of fifteen action categories, to visually analyze their classification results. We selected the following action categories: Blowing Candles, Blow Dry Hair, Cutting In Kitchen, Apply Eye Makeup, Baby Crawling, Pull Ups, Haircut, Head Massage, Punch, Playing Guitar, Playing Piano, Playing Violin, Basketball, Basketball Dunk, Biking. We keep two modules, segment scale and observed global scale, and only modify and retrain the last classification layer. The confusion matrix of the results of 15 actions at progress level of 20% is shown in Fig. 5. It can be seen intuitively from the figure that our model still has stable prediction performance for action prediction in different scenarios, even in the very early stage of actions. Only a few actions (Haircut, Blow Dry Hair, and Head Massage) with very similar external features were mispredicted. As shown in Fig6, it is an appearance comparison of Haircut, Blow Dry Hair, and Head Massage. It can be seen that three actions are difficult to distinguish, resulting in the problem of mispredicted.

TABLE V
THE ACCURACY (%) ON UCF101 DATASET UNDER SEVERAL HYPERPARAMETERS. (NOTE: LIMITED BY RESOURCES AND TIME, OUR EXPERIMENTAL RESULTS DO NOT GUARANTEE THAT ALL HYPERPARAMETERS HAVE BEEN ADJUSTED TO THE OPTIMUM.)

| Hyperparameter variables | | Observation Ratios | | | | | |
|---|---|---|---|---|---|---|---|
| | | 0.1 | 0.3 | 0.5 | 0.7 | 0.9 | Avg. |
| Hidden size of LSTM | 512 | **90.05** | **90.82** | **92.60** | **92.22** | **92.48** | **91.78** |
| | 1024 | 88.77 | 90.05 | 90.82 | 91.07 | 91.20 | 90.60 |
| | 2048 | 88.23 | 88.93 | 89.95 | 91.03 | 91.73 | 90.14 |
| Learning rate | 0.0001 | 82.14 | 84.06 | 85.97 | 87.12 | 88.01 | 85.85 |
| | 0.0005 | **90.05** | **90.82** | **92.60** | **92.22** | **92.48** | **91.78** |
| | 0.001 | 89.41 | 90.31 | 91.07 | 90.82 | 90.56 | 90.57 |
| Decay step (decay rate=0.1) | 20, 80 | 89.41 | 90.05 | 91.58 | 91.84 | 91.96 | 91.09 |
| | 40, 100 | 90.31 | 90.18 | 91.45 | 92.09 | 92.09 | 91.28 |
| | 60, 100 | **90.18** | **91.07** | **91.71** | **92.35** | **92.61** | **91.78** |

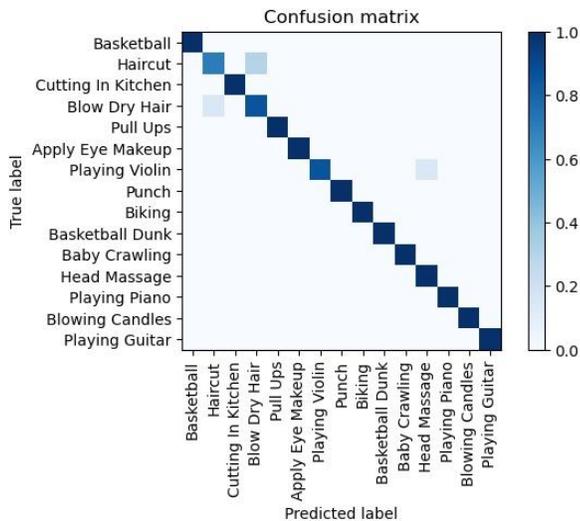

**Fig. 5.** Confusion matrix of the result of 15 classes at progress level of 20% on UCF101 dataset.

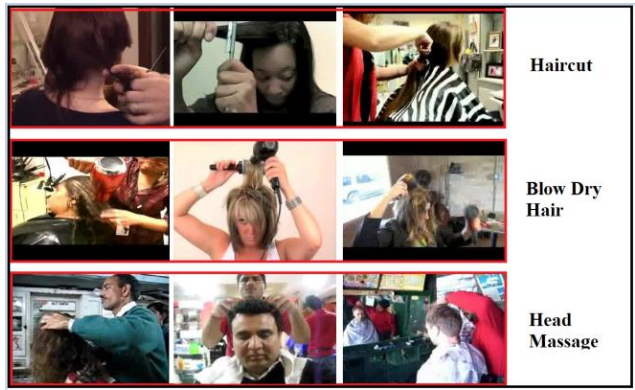

**Fig. 6.** Appearance comparison of Haircut, Blow Dry Hair, and Head Massage.

## V. CONCLUSION

In this paper, we have proposed a network model, E2EMSNet, for action prediction in videos. We propose two temporal scales, segment scale and observed global scale, to model the evolution of actions, and fuse the two scales into an end-to-end framework. A stack of 2D convolutional layers with input of RGB difference is introduced to model the local evolution of actions in a more fine-grained way. Next, the LSTM layer fuses each segment scale in the temporal dimension into an observed global scale to model the long-term evolution of actions. After experimental validation and analysis, our method possesses powerful local scale modeling capability to model ongoing actions. However, due to the growth of the time scale and the increasing noise, our observed scale cannot achieve the global modeling ability we expected for the evolving actions, which will also be the focus of our future work.

# References

[1] Liu J, Shahroudy A, Wang G, et al. Skeleton-based online action prediction using scale selection network[J]. IEEE transactions on pattern analysis and machine intelligence, 2019, 42(6): 1453-1467.

[2] Y. Hou, Z. Li, P. Wang and W. Li, "Skeleton Optical Spectra-Based Action Recognition Using Convolutional Neural Networks," in IEEE Transactions on Circuits and Systems for Video Technology, vol. 28, no. 3, pp. 807-811, 2016.

[3] H. Luo, G. Lin, Y. Yao, Z. Tang, Q. Wu and X. Hua, "Dense Semantics-Assisted Networks for Video Action Recognition," in IEEE Transactions on Circuits and Systems for Video Technology, vol. 32, no. 5, pp. 3073-3084, 2021.

[4] Feichtenhofer C, Fan H, Malik J, et al. Slowfast networks for video recognition[C]//Proceedings of the IEEE/CVF international conference on computer vision. 2019: 6202-6211.

[5] Kong Y, Tao Z, Fu Y. Deep sequential context networks for action prediction[C]//Proceedings of the IEEE conference on computer vision and pattern recognition. 2017: 1473-1481.

[6] Li M, Chen L, Lu J, et al. Order-Constrained Representation Learning for Instructional Video Prediction[J]. IEEE Transactions on Circuits and Systems for Video Technology, vol. 32, no. 8, pp. 5438-5452, 2022.

[7] Cai Y, Li H, Hu J F, et al. Action knowledge transfer for action prediction with partial videos[C]//Procee-dings of the AAAI conference on artificial intelligence. 2019, 33(01): 8118-8125.

[8] Chen L, Lu J, Song Z, et al. Recurrent semantic preserving generation for action prediction[J]. IEEE Transactions on Circuits and Systems for Video Technology, 2020, 31(1): 231-245.

[9] Ryoo M S. Human activity prediction: Early recognition of ongoing activities from streaming videos[C]//2011 International Conference on Computer Vision. IEEE, 2011: 1036-1043.

[10] Kong Y, Gao S, Sun B, et al. Action prediction from videos via memorizing hard-to-predict samples[C]//Proceedings of the AAAI Conference on Artificial Intelligence. 2018, 32(1).

[11] Wang X, Hu J F, Lai J H, et al. Progressive teacher-student learning for early action prediction[C]//Proceedings of the IEEE/CVF Conference on Computer Vision and Pattern Recognition. 2019: 3556-3565.

[12] Zhao H, Wildes R P. Spatiotemporal feature residual propagation for action prediction[C]//Proceedings of the IEEE/CVF International Conference on Computer Vision. 2019: 7003-7012.

[13] Wang L, Tong Z, Ji B, et al. Tdn: Temporal difference networks for efficient action recognition[C]//Proceedings of the IEEE/CVF Conference on Computer Vision and Pattern Recognition. 2021: 1895-1904.

[14] Lin J, Gan C, Han S. Tsm: Temporal shift module for efficient video understanding[C]//Proceedings of the IEEE/CVF International Conference on Computer Vision. 2019: 7083-7093.

[15] Simonyan K, Zisserman A. Two-stream convolution-al networks for action recognition in videos[J]. Advances in neural information processing systems, 2014, 27.

[16] Wang X, Girshick R, Gupta A, et al. Non-local neural networks[C]//Proceedings of the IEEE conference on computer vision and pattern recognition. 2018: 7794-7803.

[17] Jiang B, Wang M M, Gan W, et al. Stm: Spatiotemporal and motion encoding for action recognition[C]//Proceedings of the IEEE/CVF International Conference on Computer Vision. 2019: 2000-2009.

[18] Wang H, Tran D, Torresani L, et al. Video modeling with correlation networks[C]//Proceedings of the IEEE/CVF Conference on Computer Vision and Pattern Recognition. 2020: 352-361.

[19] Ji S, Xu W, Yang M, et al. 3D convolutional neural networks for human action recognition[J]. IEEE transactions on pattern analysis and machine intelligence, 2012, 35(1): 221-231.

[20] Tran D, Bourdev L, Fergus R, et al. Learning spatiotemporal features with 3d convolutional networks [C]//Proceedings of the IEEE international conference on computer vision. 2015: 4489-4497.

[21] Tran D, Wang H, Torresani L, et al. A closer look at

spatiotemporal convolutions for action recognition[C]// Proceedings of the IEEE conference on Computer Vision and Pattern Recognition. 2018: 6450-6459.

[22] Xie S, Sun C, Huang J, et al. Rethinking spatiotemporal feature learning: Speed-accuracy trade-offs in video classification[C]//Proceedings of the European conference on computer vision (ECCV). 2018: 305-321.

[23] Li K, Li X, Wang Y, et al. CT-net: Channel tensorization network for video classification[J]. arXiv preprint arXiv:2106.01603, 2021.

[24] Liu Z, Wang L, Wu W, et al. Tam: Temporal adaptive module for video recognition[C]//Proceedings of the IEEE/CVF International Conference on Computer Vision. 2021: 13708-13718.

[25] Li Y, Ji B, Shi X, et al. Tea: Temporal excitation and aggregation for action recognition[C]//Proceedings of the IEEE/CVF conference on computer vision and pattern recognition. 2020: 909-918.

[26] Feichtenhofer C. X3d: Expanding architectures for efficient video recognition[C]//Proceedings of the IEEE/CVF Conference on Computer Vision and Pattern Recognition. 2020: 203-213.

[27] Zhao H, Wildes R P. Review of Video Predictive Understanding: Early Action Recognition and Future Action Prediction[J]. arXiv preprint arXiv:2107. 05140, 2021.

[28] Kong Y, Kit D, Fu Y. A discriminative model with multiple temporal scales for action prediction[C]// European conference on computer vision. Springer, Cham, 2014: 596-611.

[29] Singh G, Saha S, Sapienza M, et al. Online real-time multiple spatiotemporal action localisation and prediction[C]//Proceedings of the IEEE International Conference on Computer Vision. 2017: 3637-3646.

[30] Wu X, Wang R, Hou J, et al. Spatial–temporal relation reasoning for action prediction in videos[J]. International Journal of Computer Vision, 2021, 129(5): 1484-1505.

[31] Kong Y, Fu Y. Max-margin action prediction machine [J]. IEEE transactions on pattern analysis and machine intelligence, 2015, 38(9): 1844-1858.

[32] Fernando B, Herath S. Anticipating human actions by correlating past with the future with jaccard similarity measures[C]//Proceedings of the IEEE/CVF Conference on Computer Vision and Pattern Recognition. 2021: 13224-13233.

[33] Vondrick C, Pirsiavash H, Torralba A. Anticipating visual representations from unlabeled video[C]// Proceedings of the IEEE conference on computer vision and pattern recognition. 2016: 98-106.

[34] Shi Y, Fernando B, Hartley R. Action anticipation with rbf kernelized feature mapping rnn[C]// Proceedings of the European Conference on Computer Vision (ECCV). 2018: 301-317.

[35] Gammulle H, Denman S, Sridharan S, et al. Predicting the future: A jointly learnt model for action anticipation[C]//Proceedings of the IEEE/CVF International Conference on Computer Vision. 2019: 5562-5571.

[36] Chen J, Bao W, Kong Y. Group activity prediction with sequential relational anticipation model[C]//European Conference on Computer Vision. Springer, Cham, 2020: 581-597.

[37] Oord A, Dieleman S, Zen H, et al. Wavenet: A generative model for raw audio[J]. arXiv preprint arXiv:1609.03499, 2016.

[38] Heng Wang, Alexander Kläser, Cordelia Schmid, Liu Cheng-Lin. Action Recognition by Dense Trajectories. CVPR 2011 - IEEE Conference on Computer Vision & Pattern Recognition, Jun 2011, Colorado Springs, United States. pp.3169-3176, ff10.1109/CVPR.2011.5995407ff. ffinria-00583818f

[39] Zheng Z, An G, Ruan Q. Multi-level recurrent residual networks for action recognition[J]. arXiv preprint arXiv:1711.08238, 2017.

[40] Zhao Y, Xiong Y, Wang L, et al. Temporal action detection with structured segment networks[C]// Proceedings of the IEEE International Conference on Computer Vision. 2017: 2914-2923.

[41] Chung J, Ahn S, Bengio Y. Hierarchical multiscale recurrent neural networks[J]. arXiv preprint arXiv: 1609.01704, 2016.

[42] Wang J, Wang Z, Li J, et al. Multilevel wavelet decomposition network for interpretable time series

analysis[C]//Proceedings of the 24th ACM SIGKDD International Conference on Knowledge Discovery & Data Mining. 2018: 2437-2446.

[43] Hu H, Wang L, Qi G J. Learning to adaptively scale recurrent neural networks[C]//Proceedings of the AAAI Conference on Artificial Intelligence. 2019, 33(01): 3822-3829.

[44] Campos V, Jou B, Giró-i-Nieto X, et al. Skip rnn: Learning to skip state updates in recurrent neural networks[J]. arXiv preprint arXiv:1708.06834, 2017.

[45] Zhao Y, Xiong Y, Lin D. Recognize actions by disentangling components of dynamics[C]//Procee-dings of the IEEE Conference on Computer Vision and Pattern Recognition. 2018: 6566-6575.

[46] Kong Y, Jia Y, Fu Y. Learning human interaction by interactive phrases[C]//European conference on computer vision. Springer, Berlin, Heidelberg, 2012: 300-313.

[47] Kuehne H, Jhuang H, Garrote E, et al. HMDB: a large video database for human motion recognition[C]// 2011 International conference on computer vision. IEEE, 2011: 2556-2563.

[48] Soomro K, Zamir A R, Shah M. UCF101: A dataset of 101 human actions classes from videos in the wild[J]. arXiv preprint arXiv:1212.0402, 2012.

[49] Kong Y, Tao Z, Fu Y. Adversarial action prediction networks[J]. IEEE transactions on pattern analysis and machine intelligence, 2018, 42(3): 539-553.

[50] Liu C, Gao Y, Li Z, et al. Action Prediction Network with Auxiliary Observation Ratio Regression[C]// 2021 IEEE International Conference on Multimedia and Expo (ICME). IEEE, 2021: 1-6.

[51] Chen L, Lu J, Song Z, et al. Ambiguousness-Aware State Evolution for Action Prediction[J]. IEEE Transactions on Circuits and Systems for Video Technology, vol. 32, no. 9, pp. 6058-6072, 2022.

[52] Lai S, Zheng W S, Hu J F, et al. Global-local temporal saliency action prediction[J]. IEEE Transactions on Image Processing, 2017, 27(5): 2272-2285.

[53] Hou J, Wu X, Wang R, et al. Confidence-guided self refinement for action prediction in untrimmed videos[J]. IEEE Transactions on Image Processing, 2020, 29: 6017-6031.

[54] Wu X, Zhao J, Wang R. Anticipating Future Relations via Graph Growing for Action Prediction[C]// Proceedings of the AAAI Conference on Artificial Intelligence. 2021, 35(4): 2952-2960.

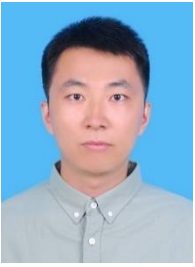

**Xiaofa Liu** received the B.S. degree from Hohai University, Nanjing, China, in 2017. He is currently pursuing the M.S. degree in mechanical engineering with the School of Modern Post, Beijing University of Posts and Telecommunications, Beijing, China. His research interests include robotics, and computer vision.

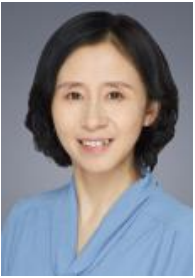

**Jianqin Yin** (Member, IEEE) received the Ph.D. degree from Shandong University, Jinan, China, in 2013. She currently is a Professor with the School of Artificial Intelligence, Beijing University of Posts and Telecommunications, Beijing, China. Her research interests include service robot, pattern recognition, machine learning, and image processing.

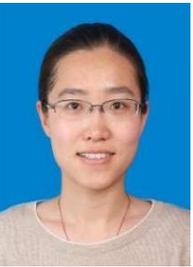

**Yuan Sun** received the Ph.D. degree from Beijing University of Aeronautics and Astronautics, Beijing, China, in 2016. She currently is an Assistant Professor with Electronic Engineering School, Beijing University of Posts and Telecommunications, Beijing, China. Her research interests include satellite navigation technology, and satellite autonomous integrity.

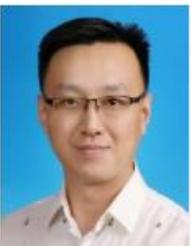

**Zhicheng Zhang** received the Ph.D. degree from Jilin University, Changchun, China, in 2011. He currently is an Associate Professor with the School of Artificial Intelligence, Beijing University of Posts and Telecommunications, Beijing, China. His research interests include Intelligent optimization and its application, signal detection and estimation, machine learning.

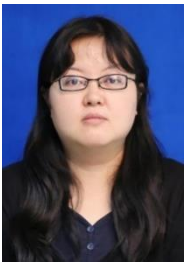

**Jin Tang** received the Ph.D. degree from Beijing Institute of Technology, Beijing, China, in 2007. currently is an Assistant Professor with Artificial Intelligence School, Beijing University of Posts and Telecommunications, Beijing, China. Her research interests include signal processing, pattern recognition, and deep learning.